\documentclass[sigconf,natbib=true]{acmart}
\AtBeginDocument{%
  }

\setcopyright{acmlicensed}
\copyrightyear{2026}
\acmYear{2026}
\setcopyright{cc}
\setcctype{by}
\acmConference[SIGIR '26]{Proceedings of the 49th International ACM SIGIR Conference on Research and Development in Information Retrieval}{July 20--24, 2026}{Melbourne, VIC, Australia}
\acmBooktitle{Proceedings of the 49th International ACM SIGIR Conference on Research and Development in Information Retrieval (SIGIR '26), July 20--24, 2026, Melbourne, VIC, Australia}
\acmDOI{10.1145/3805712.3808628}
\acmISBN{979-8-4007-2599-9/2026/07}
\usepackage{multirow}
\usepackage{pifont}
\usepackage{array}
\usepackage{hyperref}
\usepackage{balance}
\allowdisplaybreaks

\begin{document}

\title{\textsc{CommonWhy}: A Dataset for Evaluating Entity-Based Causal Commonsense Reasoning in Large Language Models}

\author{Armin Toroghi}
\authornote{Both authors contributed equally to this research.}
\affiliation{%
  \institution{University of Toronto}
  \city{Toronto, ON}
  \country{Canada}
}
\email{armin.toroghi@mail.utoronto.ca}

\author{Faeze Moradi Kalarde}
\affiliation{%
  \institution{University of Toronto}
  \city{Toronto, ON}
  \country{Canada}
}
\email{faeze.moradi@mail.utoronto.ca}
\authornotemark[1]

\author{Scott Sanner}
\affiliation{%
  \institution{University of Toronto}
  \city{Toronto, ON}
  \country{Canada}
}
\email{ssanner@mie.utoronto.ca}


\begin{abstract}
  To effectively interact with the real world, Large Language Models (LLMs) require entity-based commonsense reasoning, a challenging task that necessitates integrating factual knowledge about specific entities with commonsense inference. Existing datasets for evaluating LLM entity-based commonsense reasoning have largely focused on True/False or multiple-choice questions, leaving the explicit assessment of the model's ability in abductive reasoning about causes and effects and generating explanations largely unexamined. In this work, we introduce \textsc{CommonWhy}, a dataset of 15,000 \textit{why} questions designed to evaluate entity-based commonsense reasoning about causal relationships in LLMs. \textsc{CommonWhy} also serves as a Knowledge Graph Question Answering (KGQA) benchmark, as all supporting knowledge required to answer its queries is available in the Wikidata knowledge graph. Unlike existing KGQA datasets, which primarily test fact retrieval, \textsc{CommonWhy} targets causal commonsense reasoning, establishing a new paradigm for KGQA evaluation. Experiments with state-of-the-art LLMs and LLM-based KGQA methods reveal their significant shortcomings, including frequent factual hallucinations and failures in causal reasoning. 
\end{abstract}

\ccsdesc[500]{Information systems~Language models}
\ccsdesc[500]{Information systems~Question answering}
\ccsdesc[500]{Computing methodologies~Knowledge representation and reasoning}

\keywords{Large Language Models, Commonsense Reasoning, Knowledge Graphs, Question Answering}


\maketitle

\section{Introduction}
Large Language Models (LLMs) have ushered in a new phase of artificial intelligence, becoming increasingly useful as general-purpose assistants. Their effectiveness largely stems from their ability to encode vast amounts of factual knowledge~\citep{heinzerling2021language, ju2024large}, as well as their mastery of commonsense knowledge, the general knowledge that humans develop about the world and how it works~\citep{zhao2023large, toroghi2024right, toroghi2024verifiable}. Entity-based commonsense reasoning~\citep{creak,strategyqa,toroghi2025colota}, which involves commonsense reasoning about concrete entities such as people, places, and objects, captures both of these capabilities and is crucial for interacting with the real world. Therefore, evaluating the performance of LLMs on entity-based commonsense reasoning is essential.

To facilitate systematic assessment of entity-based commonsense reasoning, multiple benchmark datasets have been introduced~\citep{creak,strategyqa,toroghi2025colota}.
Existing datasets consist of either Yes/No questions, such as \textit{“Did Aristotle use a laptop?”}, or require the models to verify commonsense claims such as \textit{``Harry Potter can teach classes on
how to fly on a broomstick.''} by giving True/False answers, which offer only a constrained measure of the model's reasoning performance. However, none of the existing datasets evaluate LLMs’ ability to perform entity-based commonsense reasoning about causes and effects through \textit{why} questions, e.g., \textit{“Why wouldn't Katarina Barley require a visa to watch the 2024 Champions League Final in person?”}. Studying \textit{why} questions is important because it reveals whether the model can genuinely reason about causal relationships and constraints, and focuses on explicitly evaluating the model's \textit{thought} processes rather than a mere final answer which may be provided by simple pattern matching or random guesses.

In this paper, we introduce \textsc{CommonWhy}, the first dataset designed to evaluate the \textit{abductive} (explanation-driven) causal reasoning capabilities of LLMs for entity-based commonsense reasoning, comprising 15,000 carefully curated \textit{why} questions each associated with one or more answers. We find that \textsc{CommonWhy} is challenging even for the strong modern LLMs, including recent large reasoning models such as OpenAI-o3 and DeepSeek-V3.2, with models frequently failing to produce correct answers and exhibiting a high rate of hallucinations and reasoning errors. This finding is particularly striking given that these models have consistently demonstrated strong performance on existing commonsense reasoning datasets, which predominantly emphasize \textit{deductive} or verification-style reasoning~\cite{deveci2025ouroboros,guo2025benchmarking, wang2024mmlu,cotnareanu2026balanced}. Our results suggest that success on such benchmarks does not transfer to abductive causal reasoning in an entity-based setting, indicating that commonsense reasoning remains an unsolved challenge for LLMs.

\begin{figure*}
    \centering
    \includegraphics[width=0.9\linewidth]{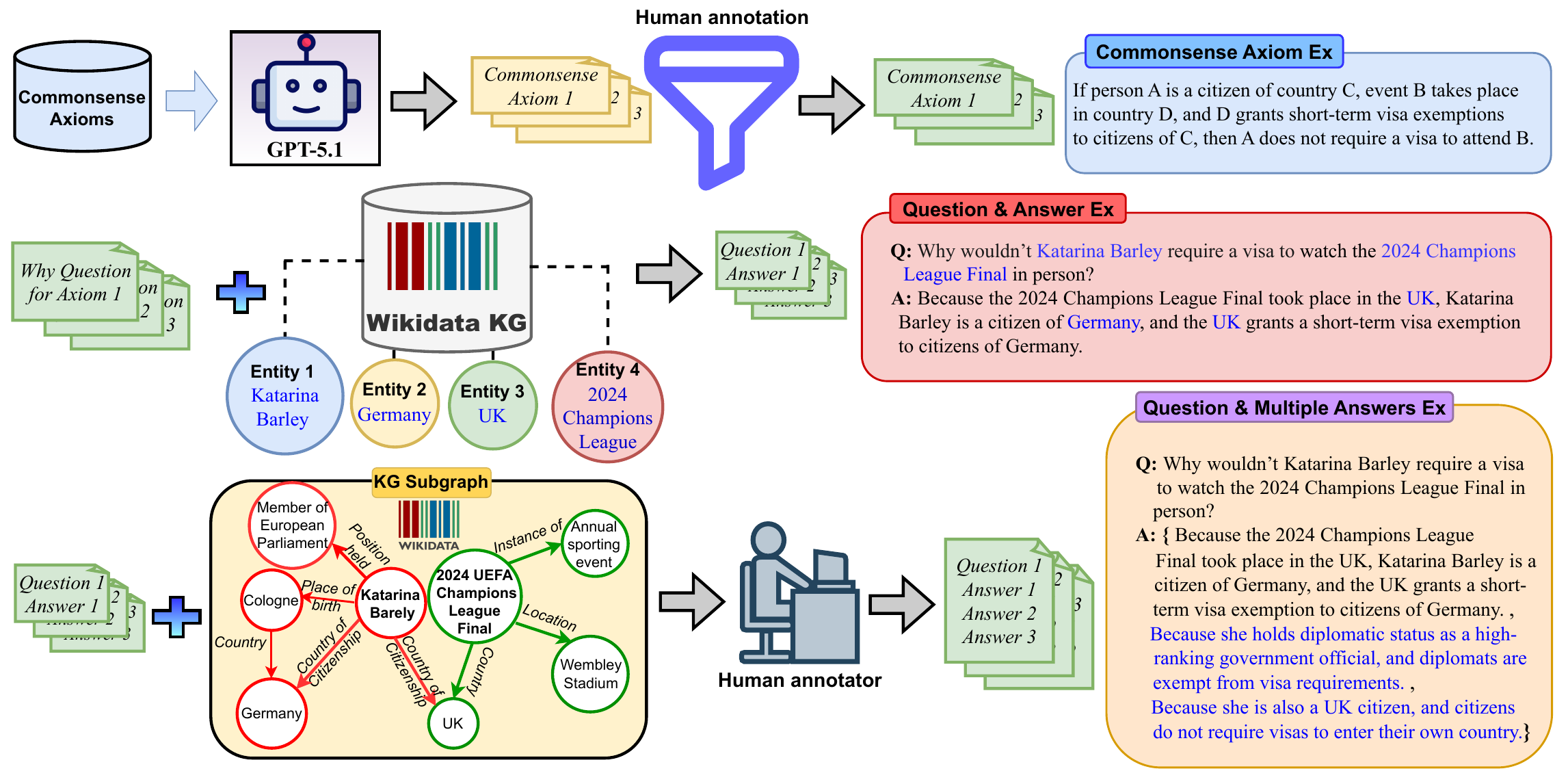}
     \caption{Overview of the \textsc{CommonWhy} dataset construction pipeline. 
    (1) Commonsense axioms extracted from existing datasets are fed to GPT-5.1 to produce additional similar commonsense axioms, after which human annotators filter out invalid ones. 
    (2) Commonsense axioms are rewritten as lifted question--answer pairs, and entities extracted from Wikidata are substituted into their variables to generate grounded pairs. 
    (3) Human annotators are provided with the generated pairs and the corresponding KG subgraphs, and are asked to identify additional plausible causes beyond the originally generated answers.}
    \label{teaserFig}
\end{figure*}

\textsc{CommonWhy} spans a broad spectrum of commonsense reasoning skills, from domain-independent capabilities such as temporal reasoning and location comparison to domain-dependent reasoning about history, professions, sports, and related real-world contexts, as shown in Figure~\ref{fig:skillsqa}. Furthermore, \textsc{CommonWhy} targets entities exclusively from the Wikidata knowledge graph (KG)~\citep{vrandevcic2014wikidata}, and the knowledge required to answer each query is ensured to be available within Wikidata. This design enables \textsc{CommonWhy} to function as a benchmark for the task of answering natural language queries over KGs, known as Knowledge Graph Question
Answering (KGQA)~\citep{yih2016value, zheng2017natural, berant2013semantic, toroghi2024bayesian}. Early KGQA datasets primarily consisted of simple questions answerable using a single KG triple~\citep{yih2016value,zhang2018variational}. More recent work has shifted toward multi-hop queries requiring compositional reasoning~\citep{grailqa, lcquad}. However, these datasets largely focus on factoid questions that can be resolved through iterative fact retrieval and entity resolution. CoLoTa~\citep{toroghi2025colota} partially departs from this trend by incorporating commonsense, but only in the form of deductive True/False queries. \textsc{CommonWhy} opens a new direction for KGQA by introducing explanation-driven abductive reasoning over KG facts that requires commonsense knowledge.

In summary, \textsc{CommonWhy} not only (i) evaluates LLMs on a type of commonsense reasoning that has previously been underexplored—causal, abductive, entity-based \textit{why} reasoning—but also (ii) opens a new avenue for KGQA research in the LLM era focused on explanation-driven reasoning over KGs, inspiring future research to develop methods capable of integrating factual knowledge with commonsense inference.

\section{Related Works}
In this section, we provide an overview of the commonsense reasoning and KGQA datasets. Table~\ref{tab:dataset_comparison} presents exemplar queries, key properties, and a comparison with \textsc{CommonWhy}.

\renewcommand{\arraystretch}{1.5}
\begin{table*}[t]
\centering
\caption{Comparison of existing commonsense reasoning and KGQA datasets with \textsc{CommonWhy}. \textsc{CommonWhy} is the first dataset that focuses on abductive causal commonsense reasoning and supports the evaluation of both LLMs and KGQA methods.}
\label{tab:dataset_comparison}
\resizebox{\textwidth}{!}{
\begin{tabular}{|
c|
c|
p{7cm}|
c|c|c|c|c|
}
\hline
\textbf{Dataset Name}
& \textbf{Category}
& \textbf{Exemplar Query} 
& \textbf{Commonsense} 
& \textbf{Factual} 
& \textbf{Usable for KGQA} 
& \textbf{Abductive} 
& \textbf{Answer Format} \\
\hline

CommonsenseQA~\citep{commonsenseqa} 
& Concept-based Commonsense 
& \textit{Where would I not want a fox? \textbf{Answer:} hen house} 
& \ding{51} & \ding{55} & \ding{55} & \ding{55} & Multiple-choice \\
\hline

PIQA~\citep{bisk2020piqa} 
& Concept-based Commonsense
& \textit{How do I find something I lost on the carpet? \textbf{Answer:} Put a hair net on the end of your vacuum and turn it on.} 
& \ding{51} & \ding{55} & \ding{55} & \ding{55} & Multiple-choice \\
\hline

StrategyQA~\citep{strategyqa} 
& Entity-based Commonsense
& \textit{Did Aristotle use a laptop? \textbf{Answer:} No.} 
& \ding{51} & \ding{51} & \ding{55} & \ding{55} & True/False \\
\hline

CREAK~\citep{creak} 
& Entity-based Commonsense
& \textit{All nuns act in holy ways. \textbf{Answer:} True.} 
& \ding{51} & \ding{51} & \ding{55} & \ding{55} & True/False \\
\hline

WebQuestions~\citep{webquestions} 
& KGQA
& \textit{What country is the Grand Bahama Island in? \textbf{Answer:} Bahamas} 
& \ding{55} & \ding{51} & \ding{51} & \ding{55} & Entity Retrieval \\
\hline

LC-QuAD~\citep{lcquad} 
& KGQA
& \textit{In which state is the Channel district? \textbf{Answer:} Florida} 
& \ding{55} & \ding{51} & \ding{51} & \ding{55} & Entity Retrieval \\
\hline

WikiWhy~\citep{ho2022wikiwhy} 
& General QA 
& \textit{Why did thousands of people evacuate from Northeastern China? 
\textbf{Answer:} The flooding triggered by Typhoon Bovalen.} 
& \ding{55}\footnotemark & \ding{51} & \ding{55} & \ding{51} & Explanation (Single-answer) \\
\hline

\textsc{CommonWhy} (Ours)
& Entity-based
Commonsense, KGQA
& \textit{Why couldn't An Chang-ok have voted for Choi Kyu-hah to become president of South Korea? \textbf{Answer:} \{``An Chang-ok is not a citizen of South Korea.'', ``An Chang-ok was born in 2003, long after Choi Kyu-hah became president in 1979.'', ``Choi Kyu-hah did not become president through a public election.''\}} 
& \ding{51} & \ding{51} & \ding{51} & \ding{51} & Explanation (Multi-answer) \\
\hline

\end{tabular}
}
\end{table*}

\subsection{Commonsense Reasoning Datasets}
Humans naturally develop an understanding 
of how the world works, called \emph{commonsense knowledge}, and the ability to reason with it, known as \emph{commonsense reasoning}~\citep{liu2004conceptnet}. To interact with the real world, AI agents must also possess this knowledge and reasoning ability~\citep{moore1982role, liu2004conceptnet, davis2015commonsense,baroni2017commai, toroghi2025llm}. 
To evaluate AI systems’ commonsense reasoning, several datasets have been developed. 
These datasets are divided into two categories: (i) concept-based and (ii) entity-based, which we summarize below.


\sloppy
\noindent
\textbf{(i) Concept-based Datasets} evaluate commonsense reasoning about general concepts or hypothetical scenarios. CommonsenseQA~\cite{commonsenseqa} is a multiple-choice benchmark in which crowd workers generate questions designed to distinguish target entities associated with a source concept from ConceptNet~\cite{concept-net}. ATOMIC~\citep{sap2019atomic} provides an atlas of causal commonsense rules in \textit{if-then} form. Some of these datasets also target specific reasoning skills: PIQA~\citep{bisk2020piqa} focuses on physical commonsense by asking which of two actions plausibly achieves a given goal, while SocialIQA~\citep{sap2019social} explores social commonsense, querying motivations, consequences, and emotional reactions in hypothetical social scenarios. Another group of works in this paradigm, such as COPA~\cite{roemmele2011choice} and its variants~\cite{brassard2022copa,ponti2020xcopa}, target commonsense causal reasoning by presenting a premise with two candidate alternatives, asking the model to select the more plausible cause or effect.

\noindent
\textbf{(ii) Entity-based Datasets} evaluate grounded commonsense reasoning about specific objects, events, or people, requiring models to combine commonsense with factual knowledge. StrategyQA~\citep{strategyqa} consists of Yes/No questions where reasoning steps are implicit and must be inferred using a strategy, with each query annotated with evidence from Wikipedia. Crowd workers generate questions given an entity term, and adversarial models are used to iteratively create harder queries. CREAK~\citep{creak} similarly contains claims about entities labeled True or False, also written by crowd workers using Wikipedia, but without a model-in-the-loop. While this human-based approach produces diverse and challenging questions, it tends to focus on popular entities, which are well-represented in the training data of LLMs. CoLoTa~\citep{toroghi2025colota} replaces these popular entities in StrategyQA and CREAK with obscure ones to study the influence of entity popularity on the performance.

Despite the diversity of existing commonsense reasoning datasets, none of them evaluate abductive causal reasoning over entities in a way that requires explicit explanation; \textsc{CommonWhy} fills this gap by introducing \textit{why} questions to entity-based commonsense reasoning. The closest existing dataset to \textsc{CommonWhy} is WikiWhy~\cite{ho2022wikiwhy}, which focuses on answering \textit{why} questions but does not target commonsense reasoning as required for real-world decision-making. Most WikiWhy questions can be answered by extracting or paraphrasing explanations explicitly stated in Wikipedia articles, raising the risk that LLMs rely on memorization rather than genuine reasoning
~\cite{xie2025memorization,berglundreversal}. In contrast, \textsc{CommonWhy} targets entity-based commonsense reasoning in settings where the relevant reasoning is not explicitly stated, requiring models to combine factual knowledge with implicit commonsense assumptions.

\footnotetext{{Although some queries require commonsense reasoning, WikiWhy questions are extracted from Wikipedia articles and most queries can be answered verbatim.}}

\subsection{KGQA Datasets}

KGs~\citep{hogan2021knowledge} are widely used to represent relational information about entities, with applications ranging from healthcare~\citep{rastogi2020personal} to recommendation systems~\citep{raza2024comprehensive, toroghi2023bayesian, tang2023logicrec}. Retrieving information from KGs has traditionally relied on query languages such as RQL~\cite{karvounarakis2002rql} and SPARQL~\citep{seaborne2008sparql}, which require technical expertise and limit accessibility~\citep{toroghi2024right}. To address this, KGQA has emerged as a means of answering natural language queries over KGs~\citep{yih2016value, zheng2017natural, berant2013semantic, toroghi2024bayesian, guo2024cr}.

Early KGQA datasets focused on simple questions answerable with a single KG triple, while later benchmarks introduced multi-hop reasoning over larger graphs. Representative datasets include WebQuestions~\cite{webquestions} and its SPARQL extension, WebQuestionsSP~\cite{webquestions-sp} and LC-QuAD~\cite{lcquad} grounded in DBpedia~\cite{lehmann2015dbpedia}, and GrailQA~\cite{grailqa} based on Freebase~\cite{freebase}. Despite increased structural complexity, these datasets are predominantly constructed from executable logical forms and therefore focus on factoid questions, limiting their ability to evaluate commonsense reasoning beyond iterative fact retrieval. Although CoLoTa~\cite{toroghi2025colota} introduces commonsense into KGQA, it remains limited to deductive True/False reasoning. \textsc{CommonWhy} advances the task by evaluating abductive causal reasoning through explanation-based answers to natural language \textit{why} questions.
\begin{figure}
  \centering
  \includegraphics[
    width=1\linewidth
  ]{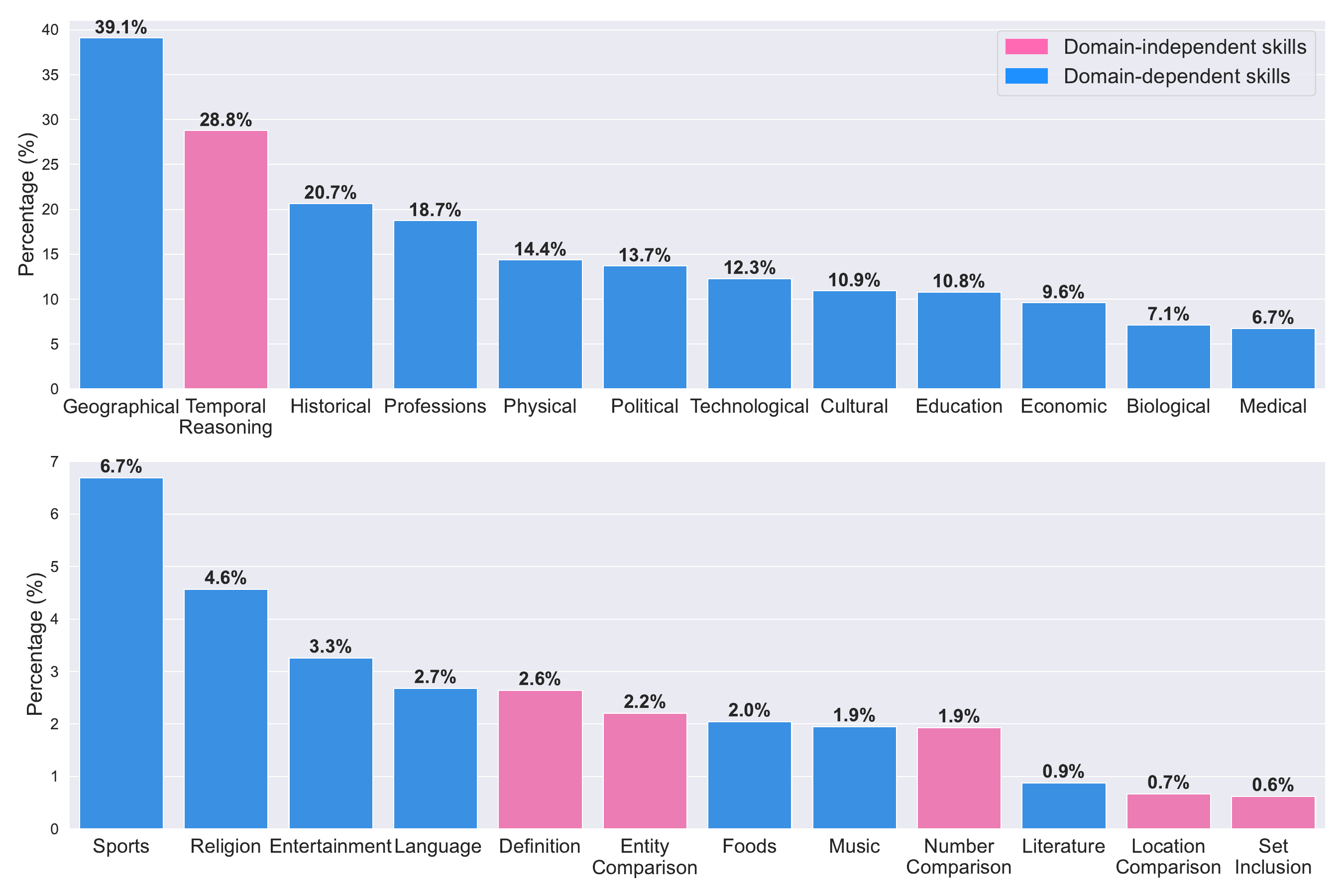}
  \caption{Distribution of reasoning skills in \textsc{CommonWhy}.} 
  \label{fig:skillsqa}
\end{figure}

\section{\textsc{CommonWhy} Dataset}

In this section, we first describe the structure and properties of the \textsc{CommonWhy} dataset~\footnote{The dataset and code are available \href{https://github.com/faezemoradik/CommonWhyDataset.git}{here}.}, and next present the methodology used for its construction.

\subsection{Description}
\textsc{CommonWhy} comprises 15,000 questions each associated with one or more answers centered on entities from the Wikidata KG, designed to evaluate abductive causal commonsense reasoning. 
Queries of \textsc{CommonWhy} cover a diverse range of commonsense reasoning skills as shown in Figure~\ref{fig:skillsqa}. Each dataset entry consists of: (i) a \textit{why} question that targets at least one Wikidata entity; (ii) the unique Wikidata identifier (QID) of the anchor entities mentioned in the query; (iii) a relevant Wikidata subgraph containing the factual information required to answer the question; (iv) a general commonsense axiom, expressed in natural language, that captures the implicit knowledge underlying the causal relationship between the question and its answers; and (v) one or more ground-truth answers that specify possible causes for the effect described in the query. We describe each constituent in detail below.

\noindent
\textbf{(i) Query.}
A \textsc{CommonWhy} query is an interrogative sentence beginning with \textit{why} that asks for the cause of an effect described in the question. Each query $q$ is associated with $L \geq 1$ ground-truth answers $\{a_q^{(l)}\}_{l=1}^{L}$, which can be derived by combining knowledge graph facts with abductive causal commonsense reasoning.

\noindent 
\textbf{(ii) Wikidata entities.} 
Each query refers to a set of anchor entities drawn from Wikidata, for which factual information is required to answer the question. For each anchor entity, we provide both its label and QID, which can be used to retrieve the required facts.

\noindent
\textbf{(iii) Relevant KG Subgraph.} Queries in \textsc{CommonWhy} are curated to ensure that all factual knowledge required to answer them is contained within the Wikidata KG. For each query, we provide the set of relevant facts extracted from a Wikidata subgraph, represented as triples of the form $(h,r,t)$, in which $h$, $r$, and $t$ denote the head entity, relation, and tail entity, respectively. These triples collectively supply the factual evidence needed to support the answer. For triples containing qualifiers in the format \textit{relation-entity} pairs that provide additional context to the original KG triple, we represent them as hyper-relational triples $\big((h, r, t), \{(k_i, v_i)\}_{i=1}^m\big)$, where each $(k_i,v_i)$ corresponds to a qualifier relation and its associated entity. 


\noindent
\textbf{(iv) Commonsense Axiom.} The commonsense axiom is a logical statement, expressed in natural language, that formally captures the causal commonsense knowledge that links an observed effect stated in the query to its plausible underlying causes. Specifically, it formalizes the properties of the entities mentioned in the query and the relations among them as premises that, when satisfied, constitute a sufficient causal explanation for the effect.

Formally, let $\mathcal{E}_q = \{ e_{1,q}, \ldots, e_{|\mathcal{E}_q|,q} \}$ denote the set of entities referenced in a query $q$. The commonsense axiom $I_q$ is a natural-language realization of a First-Order Logic (FOL) expression
\begin{equation}
a_q \;\land\;
\left( \bigwedge\limits_{i=1}^{|\mathcal{P}|} \bigwedge\limits_{j=1}^{|\mathcal{E}_q|} P_i(e_{j,q}) \right)
\;\land\;
\left( \bigwedge\limits_{i=1}^{|\mathcal{F}|} \bigwedge\limits_{j=1}^{|\mathcal{E}_q|} 
F_i(e_{j,q})\, \langle \mathit{op}_j^i \rangle\, e_{j,q}^i \right)
\;\Rightarrow\; \mathit{effect}_q, \nonumber
\end{equation}
where $\mathcal{P} = \{ P_1, \ldots, P_{|\mathcal{P}|} \}$ is a set of predicates, $\mathcal{F} = \{ F_1, \ldots, F_{|\mathcal{F}|} \}$ is a set of functions, $\langle \mathit{op}_j^i \rangle \in \{ =, \neq, <, \leq, >, \geq \}$ denotes a (dis)equality or comparison operator, and $a_q$ represents a candidate cause.

Under this formulation, answering a query amounts to abductively identifying one or more causes $a_q$ such that, together with the factual knowledge grounded in the KG, the commonsense axiom implies the effect described in the query.

\noindent
\textbf{(v) Answers.} In \textsc{CommonWhy}, answers correspond to plausible causes that explain the effect described in a query. Since real-world effects may admit multiple valid causal explanations, the dataset adopts a multi-answer formulation in which each query $q$ is associated with a set of correct answers $\{a_q^{(l)}\}_{l=1}^{L}$, with $L \geq 1$. Each answer represents a distinct abductive hypothesis that, when combined with factual knowledge from the KG and relevant commonsense assumptions, is sufficient to entail the observed effect. For example, the query \textit{``Why wouldn’t Katarina Barley need a visa to travel to the UK?''} admits multiple valid causes, such as: \textit{because she is a German citizen and German citizens do not require visas to travel to the UK}; \textit{because she holds diplomatic status as a high-ranking government official and diplomats are exempt from visa requirements}; or \textit{because she is also a UK citizen and citizens do not require visas to enter their own country}. Rather than enforcing a single explanation, \textsc{CommonWhy} explicitly models these alternative causes as separate but equally valid ground-truth answers, reflecting the inherently non-unique and abductive nature of commonsense causal reasoning.

\subsection{Methodology}
The overall framework for constructing the \textsc{CommonWhy} dataset consists of three steps and is illustrated in Figure~\ref{teaserFig}. Below, we describe each step in detail.

\subsubsection{Axiom Generation}Constructing why questions that genuinely require commonsense reasoning is challenging. The only existing large-scale dataset of why questions widely used in LLM evaluation, WikiWhy~\cite{ho2022wikiwhy}, generates queries by extracting sentences from Wikipedia that contain explicit causal markers such as \textit{because} or \textit{due to}. While scalable, this approach produces questions that are often answerable by directly retrieving or paraphrasing the source text, and therefore primarily evaluates retrieval rather than implicit commonsense reasoning. Since \textsc{CommonWhy} targets abductive causal reasoning grounded in unstated assumptions, this extraction-based strategy is insufficient. Moreover, building commonsense reasoning datasets on a scale has traditionally required substantial human effort~\cite{ismayilzada2023crow,maharana2022grada}. However, recent works show that LLMs have become increasingly effective in learning and generalizing lifted commonsense axioms, with near-ideal performance in existing concept-based benchmarks~\citep{guo2025benchmarking, wang2024mmlu,cotnareanu2026balanced}. 
This motivates us to leverage LLMs’ capability to generate lifted commonsense axioms. For entity-grounded query generation introduced in the subsequent step, however, we rely on a KG rather than LLMs, as LLMs are prone to hallucinations and factual inaccuracies.

Concretely, we provide GPT-5.1 with a set of commonsense axioms manually extracted from the StrategyQA~\citep{strategyqa}, CREAK~\citep{creak}, and CoLoTa~\cite{toroghi2025colota} datasets, and use them as few-shot examples. The model is then prompted to creatively expand this set by proposing additional axioms that follow similar causal structures. For example, given an exemplar axiom such as \textit{``If person A is from country C, and museum B is also in country C, then A does not require a visa to visit B''}, the model generates related axioms like \textit{``If person A is a citizen of country C, event B takes place in country D, and D grants short-term visa exemptions to citizens of C, then A does not require a visa to attend B''}. To ensure quality and correctness, all LLM-generated axioms are independently reviewed by two human annotators, who verify whether each rule aligns with human commonsense. We retain only axioms unanimously judged to be valid. 


\subsubsection{Query Generation}
After generating a diverse set of valid commonsense axioms, each axiom is rewritten into a lifted question–answer pair.
Next, we ground these question-answer pairs by instantiating their abstract variables with concrete entities from Wikidata. Specifically, for each lifted pair, we identify entities that satisfy the specified properties and relations. For example, given the lifted pair \textit{“Why wouldn’t Person A require a visa to attend event B?”} with the answer \textit{“Because event B takes place in country D, Person A is a citizen of country C, and D grants short-term visa exemptions to citizens of C”}, we must instantiate the variables $A$, $B$, $C$, and $D$ with real-world entities that fulfill these constraints. To this end, we construct SPARQL~\citep{seaborne2008sparql} queries over Wikidata to retrieve candidate persons, events, and countries that jointly satisfy the required factual conditions, and extract the relevant subgraph containing the supporting triples. This grounding process results in fully instantiated, entity-specific queries paired with at least one valid answer that can only be obtained by combining factual knowledge from the KG with abductive causal commonsense reasoning.

\renewcommand{\arraystretch}{1.5}
\begin{table*}[t]
\centering
\caption{Performance comparison on Long-Tail and Head subsets of \textsc{CommonWhy} across different LLMs and KGQA methods.}
\label{tab:longtail_head_results}
\resizebox{\textwidth}{!}{
\begin{tabular}{|
l|
c|c|c|c|c|c|c|
c|c|c|c|c|c|c|
}
\hline
\multirow{2}{*}{\parbox[c]{0.5cm}{\centering Model}}
& \multicolumn{7}{c|}{Long-Tail}
& \multicolumn{7}{c|}{Head} \\
\cline{2-15}
& BS-Prec & BS-Rec & BS-F1 & ROUGE-L & BLEU (\%) & METEOR & Correctness ($\%$)
& BS-Prec & BS-Rec & BS-F1 & ROUGE-L & BLEU (\%) & METEOR & Correctness ($\%$) \\
\hline
GPT-5.1~\cite{openai2025gpt51} & 0.072 & 0.424 & 0.242 & 0.223 & 4.92 & 0.327 & 57.09 & 0.063 & 0.426 & 0.239 & 0.219 & 4.56 & 0.319  & 60.49  \\
\hline
Gemini-2.5-Flash~\cite{comanici2025gemini} & \textbf{0.277} & \textbf{0.476} & \textbf{0.375} & \textbf{0.318} & \textbf{9.23} & \textbf{0.359} & 43.83 & \textbf{0.274} & \textbf{0.471} & \textbf{0.371} & \textbf{0.315} & \textbf{9.13} & \textbf{0.358} & 46.21 \\
\hline
Llama 3.3-70B~\cite{meta2024llama33_70b} & 0.214 & 0.435 & 0.323 & 0.257 & 7.26 & 0.327 & 39.20 & 0.221 & 0.437 & 0.328 & 0.258 & 7.03 & 0.324 & 43.43 \\
\hline
DeepSeek-V3.2~\cite{deepseek2025v32} & 0.032 & 0.394 & 0.207 & 0.188 & 3.72 & 0.274 & 40.89 & 0.026 & 0.397 & 0.205 & 0.185 & 3.62 & 0.273 & 41.60 \\
\hline
OpenAI-o3~\cite{openai2025o3} & 0.172 & 0.412 & 0.289 & 0.258 & 6.11 & 0.309 & \textbf{67.93} & 0.161 & 0.417 & 0.286 & 0.254 & 5.78 & 0.309 & \textbf{68.51} \\
\hline
\hline
KB-Binder~\cite{kb-binder} & 0.009 & 0.122 & 0.042 & 0.099 & 0.911 & 0.145 & 18.48 & 0.014 & 0.181 & 0.083 & 0.096 & 1.88 & 0.156 & 20.45 \\
\hline
KGR~\cite{kgr} & 0.092 & 0.403 & 0.231 & 0.230 & 5.02 & 0.289 & 56.22 & 0.083 & 0.448 & 0.251 & 0.220 & 4.89 & 0.323 & 62.33 \\
\hline
\end{tabular}
}
\end{table*}

The work in \cite{why-would-you-ask-it-that-way} has shown that questions in many QA datasets are often phrased unnaturally, and that model performance degrades significantly when evaluated on more natural reformulations. To avoid this issue, we follow their methodology and explicitly rewrite the question answers into fluent, human-like why questions, ensuring that \textsc{CommonWhy} evaluates reasoning ability rather than robustness to unnatural question phrasing.

\subsubsection{Multi-Answer Annotation}
Once queries are instantiated and grounded in Wikidata, we complete the set of plausible answers by identifying all valid causes that could explain the effect stated in the query, given the retrieved KG subgraph. Importantly, this set of causes is entity-dependent: while a lifted commonsense axiom specifies a general causal pattern, the concrete properties of the instantiated entities may introduce additional valid answers. For example, although a generic German citizen may not require a visa to attend the \textit{2024 Champions League Final} in London solely due to their nationality, a specific individual such as \textit{Katarina Barley} admits further valid causes, like holding diplomatic status or possessing UK citizenship. To capture this variability, two human annotators independently inspect the KG subgraphs associated with each instantiated query and enumerate all plausible causal explanations supported by the available facts and commonsense assumptions. Only those causes identified by both annotators are retained, ensuring high precision and consistency. This process yields a set of answers that reflects real-world causal reasoning, where outcomes often admit multiple, simultaneously valid explanations.

\begin{figure}
    \centering
    \includegraphics[width=0.99\linewidth]{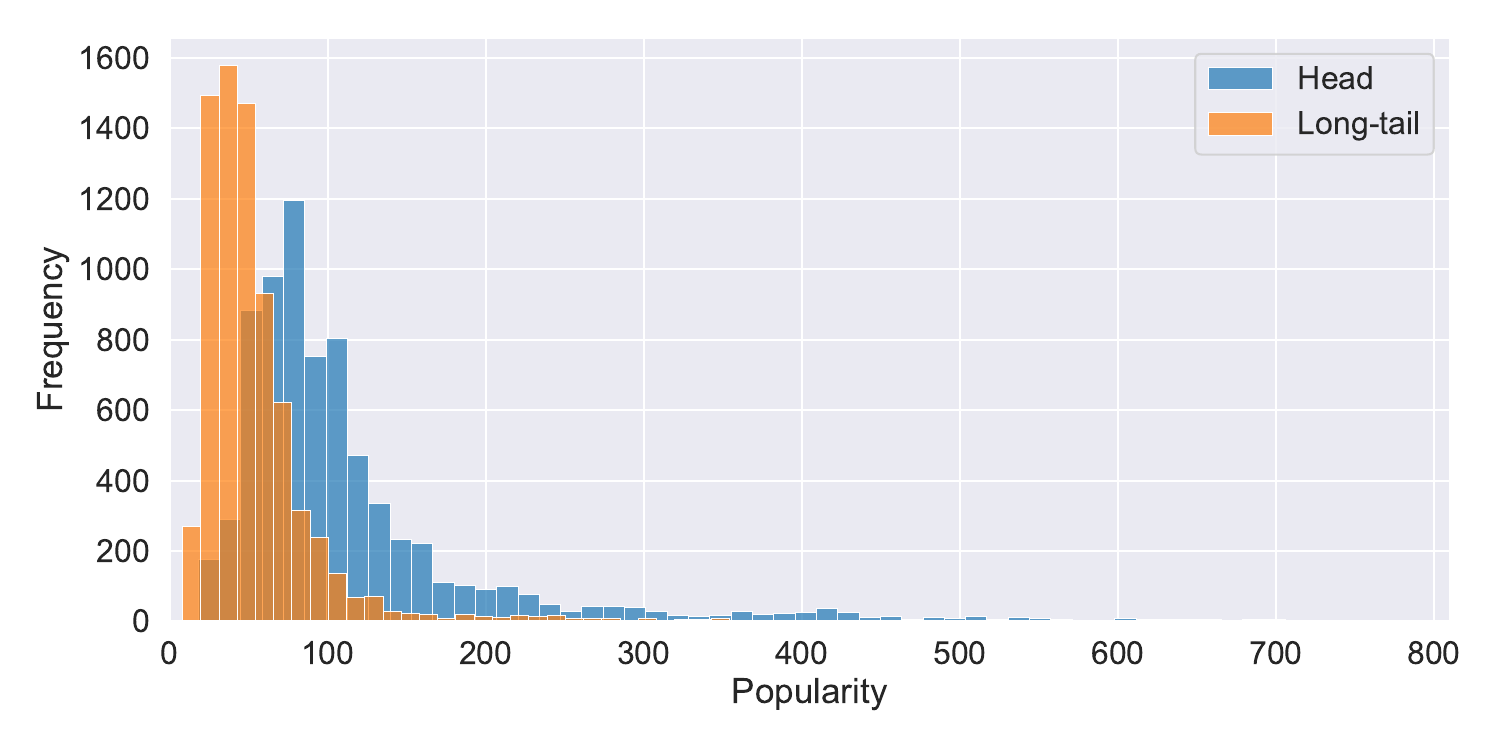}
    \caption{Entity popularities in the head and long-tail splits.}
    \label{fig:popularities}
\end{figure}

\begin{figure}[]
  \centering
  \includegraphics[
    width=0.99\linewidth]{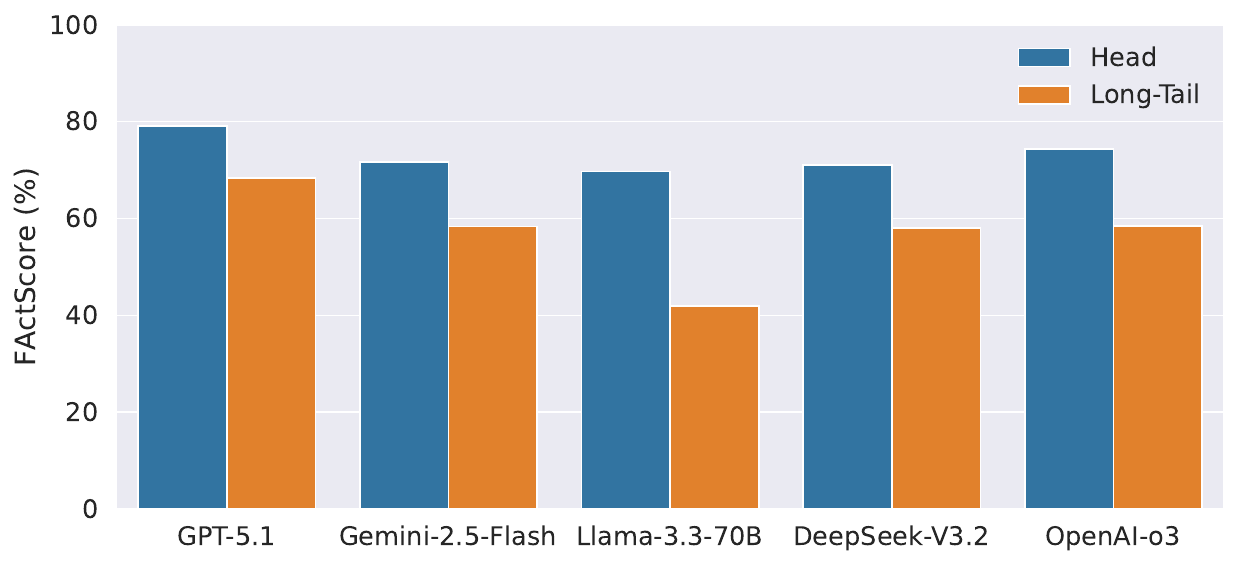}
  \caption{FActScores obtained by different LLMs across head and long-tail splits.}
  \label{fig:factscore}
\end{figure}

\section{Experiments}

\label{sec:experiments}
\subsection{Baselines}
\textsc{CommonWhy} is designed to function as both an entity-based commonsense reasoning dataset for evaluating LLMs, as well as a KGQA dataset. Hence, we evaluate it on both LLMs and KGQA methods.

\noindent \textbf{LLM Baselines.} We evaluate widely-used and modern LLMs: GPT-5.1~\cite{openai2025gpt51}, Gemini-2.5-Flash~\cite{comanici2025gemini}, Llama-3.3-70B~\cite{meta2024llama33_70b}, and two prominent reasoning models: OpenAI-o3~\cite{openai2025o3}, and DeepSeek-V3.2~\cite{liu2025deepseek, deepseek2025v32}.

\noindent \textbf{KGQA Baselines.}
Most existing KGQA methods focus on answering factoid queries by translating them into structured languages such as SPARQL~\citep{seaborne2008sparql}, and do not explicitly support commonsense reasoning. 
Hence, we evaluate \textsc{CommonWhy} using two strong LLM-based KGQA baselines that rely on LLMs to supply the required commonsense knowledge: (i) KB-Binder~\cite{kb-binder}, which employs in-context learning to translate natural language queries into SPARQL, and (ii) KGR~\cite{kgr}, which generates candidate claims with an LLM and retrofits them against the KG to enforce factual consistency. We use GPT-5.1 as the LLM backbone for both methods.

\subsection{Evaluation Metrics}
Answering \textit{why} questions in \textsc{CommonWhy} involves generating explanations that identify plausible causes of an observed effect, which differs fundamentally from the accuracy-based evaluation used in True/False or multiple-choice commonsense benchmarks. As a result, we adopt an evaluation scheme tailored to explanation-driven causal reasoning rather than discrete answer selection.

\noindent\textbf{Answer Correctness.} We evaluate whether a model’s response is logically equivalent to at least one of the ground-truth causal explanations provided for a query, regardless of surface-level phrasing differences. Because exhaustive human evaluation is costly, we use GPT-4o~\cite{openai2024gpt4o} as an LLM judge to determine logical equivalence between generated answers and any of the ground-truth answers. 

\noindent\textbf{Standard Long-form Generation Metrics.} We report standard metrics for long-form generation: BERTScore (BS)~\citep{zhangbertscore} precision, recall, and F1; ROUGE-L~\citep{lin2004rouge}; and METEOR~\citep{banerjee2005meteor}. While these metrics do not directly evaluate causal reasoning, they provide complementary signals about explanation quality and help characterize how closely model outputs align with ground-truth answers. In our multi-label setting, where each query may have multiple valid ground-truth answers, we compute each metric between the model prediction and every reference answer, and report the maximum score across references.

\noindent\textbf{Answer Factuality.} FActScore~\citep{factscore} quantifies the factual consistency of an LLM-generated answer by measuring the proportion of its atomic claims that are supported by an external knowledge source $C$. Atomic claims are validated against Wikidata and, when unsupported, cross-checked using Wikipedia. Given an answer $a_q$, consisting of the set of atomic facts $A_q$, the FActScore is computed as $f(a_q) = \frac{1}{|A_q|} \sum_{f \in A_q} \mathbb{I}\left(f \text{ is supported by } C\right)$, where $|A_q|$ denotes the cardinality of the set
$A_q$, and $\mathbb{I}(\cdot)$ is the binary indicator function that maps a true statement to 1 and a false statement to 0.


\subsection{Results}
To analyze the impact of entity popularity on performance, we split \textsc{CommonWhy} into equally-sized head and long-tail subsets based on Wikidata triple counts, following CoLoTa~\citep{toroghi2025colota}. The popularity distributions for both splits are shown in Figure~\ref{fig:popularities}. This section summarizes the main results of the experiments.

\noindent\textbf{\textsc{CommonWhy} reveals limitations of state-of-the-art LLMs in entity-based commonsense reasoning:} Results of the experiments are summarized in Table \ref{tab:longtail_head_results}. Overall, answer correctness is low across all evaluated models, underscoring the difficulty of abductive causal commonsense reasoning over entities. The best-performing model, OpenAI-o3, a prominent reasoning model, achieves correctness scores of only $67.93\%$ and $68.51\%$ on the long-tail and head splits, respectively. DeepSeek-V3.2, another well-known reasoning model, performs substantially worse, reaching $40.89\%$ and $41.60\%$ correctness on the two splits. These results highlight that even state-of-the-art reasoning models struggle with entity-grounded abductive causal reasoning.

Furthermore, we observe that standard long-form generation metrics are not reliable indicators of causal correctness. For example, Gemini-2.5-Flash consistently achieves the highest scores on BERTScore, ROUGE-L, and METEOR, yet its answer correctness remains lower than that of OpenAI-o3 and GPT-5.1. This discrepancy suggests that lexical or semantic similarity to reference explanations does not necessarily reflect correct abductive reasoning, indicating that standard long-form generation metrics are insufficient and that evaluating causal correctness requires LLM-based judges.

\noindent\textbf{Existing KGQA methods cannot perform causal commonsense reasoning to answer \textsc{CommonWhy} queries:}
Both KGQA approaches, KB-Binder~\citep{kb-binder} and KGR~\citep{kgr}, perform poorly on the abductive causal commonsense reasoning task introduced in \textsc{CommonWhy}. In particular, KB-Binder exhibits especially weak performance, as it relies on translating questions into SPARQL queries—a formulation ill-suited for answering \textit{why} questions that require implicit commonsense and causal inference beyond explicit triples in the knowledge graph. KGR likewise fails to significantly improve over directly prompting its backbone LLM (GPT-5.1), indicating that current KGQA pipelines remain inadequate for abductive causal reasoning, highlighting the need for future methodological advances.

\noindent\textbf{LLMs and KGQA methods perform worse on abductive causal commonsense reasoning over long- tail entities:} Prior works have shown that LLM performance on True/False entity-based commonsense questions declines as entity popularity decreases~\citep{toroghi2025colota}. We observe the same trend for abductive causal commonsense reasoning: Across all LLMs and KGQA methods, performance consistently declines when moving from queries involving head entities to those targeting long-tail entities. Importantly, in constructing the head and long-tail splits of \textsc{CommonWhy}, we control for reasoning complexity by ensuring that both splits are derived from the same set of commonsense axioms, such that the underlying reasoning patterns remain identical and only entity popularity varies. This design makes \textsc{CommonWhy} particularly suitable for studying and developing methods that improve the robustness of reasoning models to variations in entity popularity.

\noindent\textbf{LLMs suffer from hallucinations in causal commonsense reasoning, particularly over long-tail entities.}
To assess the factual reliability of model outputs beyond answer correctness, we compute FActScore for the responses judged as correct, following \citet{kazemi2023lambada}. As shown in Figure~\ref{fig:factscore}, all models achieve modest FActScores, indicating that even logically correct answers often contain hallucinated facts. This issue becomes more pronounced for long-tail entities. These findings highlight that \textsc{CommonWhy} not only challenges models’ abductive causal reasoning abilities, but also exposes their tendency to introduce factual hallucinations when integrating knowledge with commonsense inference.

\section{Conclusion}
We proposed \textsc{CommonWhy}, a new dataset for evaluating the abductive causal commonsense reasoning capability of LLMs in entity-based settings. We observed a high rate of hallucinations and reasoning errors even from leading reasoning LLMs on \textsc{CommonWhy}. 
Since the factual knowledge required to answer its queries is grounded in Wikidata, \textsc{CommonWhy} also establishes a new direction for KGQA: answering \textit{why} questions that require abductive causal reasoning rather than purely factual retrieval. Our experiments with two LLM-based KGQA methods further show the insufficiency of current methods for this setting. Overall, \textsc{CommonWhy} offers a challenging testbed for studying causal reasoning in LLMs, and paves the way for future reasoning-oriented KGQA research. 

\begin{acks}
This work was supported by the Institute of Information \& Communications Technology Planning \& Evaluation (IITP) grant funded by the Korean Government (MSIT) (No. RS-2024-00457882, National AI Research Lab Project).
\end{acks}

\bibliographystyle{ACM-Reference-Format}
\balance
\bibliography{sample-base}

\end{document}